\documentclass[letterpaper, 10 pt, conference]{ieeeconf}  

\IEEEoverridecommandlockouts                              

\usepackage{graphicx} 
\usepackage{etoolbox}
\usepackage{longtable}
\usepackage{booktabs}
\usepackage{caption}
\usepackage{subfig}
\graphicspath{{images/}} 
\usepackage{float}
\usepackage{cite}
\usepackage{textcomp}
\usepackage[colorlinks=true,linkcolor=black,anchorcolor=black,citecolor=black,filecolor=black,menucolor=black,runcolor=black,urlcolor=black]{hyperref} 
\usepackage{rotating}

\usepackage[dvipsnames]{xcolor}
\usepackage{tikz}
\usetikzlibrary{automata, shapes, arrows, positioning, calc, fit, backgrounds, decorations.pathreplacing}
\usepackage{amsmath,bm,times}
\usepackage{verbatim}
\usepackage{cleveref}
\usepackage[ruled]{algorithm}
\usepackage{algpseudocode}
\usepackage{paralist}

\usepackage{stfloats}
\title{\LARGE \bf

Learning and Online Replication of Grasp Forces from Electromyography Signals for Prosthetic Finger Control

}

\author{Robin Arbaud$^{*1,2}$, Elisa Motta$^{*1}$, Marco Domenico Avaro$^{3}$, Stefano Picinich$^{3}$, \\ Marta Lorenzini$^{1}$, and Arash Ajoudani$^{1}$
\thanks{This work was supported by PR FESR 2021–2027: Incentivi alle imprese per attività collaborativa di ricerca industriale e sviluppo sperimentale. Bando DGR 2026/2021.
Progetto RE-FINGER: Prat. n. 2022/38. }
\thanks{$^{*}$Equally contributed.}
\thanks{$^{1}$Human-Robot Interfaces and Interaction Laboratory, Istituto Italiano di Tecnologia, Genoa, Italy. $^{2}$Dept. of Informatics, Bioengineering, Robotics, and System Engineering, University of Genoa, Genoa, Italy. $^{3}$Airworks SRL, Povoletto, Italy.}
\thanks{Emails: {\tt\small Robin.Arbaud|Elisa.Motta@iit.it}}%
\thanks{© 2025 IEEE.  Personal use of this material is permitted.  Permission from IEEE must be obtained for all other uses, in any current or future media, including reprinting/republishing this material for advertising or promotional purposes, creating new collective works, for resale or redistribution to servers or lists, or reuse of any copyrighted component of this work in other works.}
}

\begin{document}

\maketitle
\thispagestyle{empty}
\pagestyle{empty}

\begin{abstract}

Partial hand amputations significantly affect the physical and psychosocial well-being of individuals, yet intuitive control of externally powered prostheses remains an open challenge.
To address this gap, we developed a force-controlled prosthetic finger activated by electromyography (EMG) signals. 
The prototype, constructed around a wrist brace, functions as a supernumerary finger placed near the index, allowing for early-stage evaluation on unimpaired subjects.
A neural network-based model was then implemented to estimate fingertip forces from EMG inputs, allowing for online adjustment of the prosthetic finger grip strength. 
The force estimation model was validated through experiments with ten participants, demonstrating its effectiveness in predicting forces. Additionally, online trials with four users wearing the prosthesis exhibited precise control over the device. Our findings highlight the potential of using EMG-based force estimation to enhance the functionality of prosthetic fingers.

\end{abstract}

\section{INTRODUCTION}

Upper extremity amputations make up 3\% to 23\% of all amputations, with approximately 50\% to 90\% of these being related to trauma. Specifically, 90\% of upper limb amputation cases involve partial hand loss, affecting one or more fingers \cite{giladi2013surgical}. This form of amputation can cause physical, psychosocial, and even economic damage, as several individuals are unable to return to their prior occupations. Hence, the development of prosthetic limbs offers them an opportunity to 
restore their independence and quality of life.
While there is significant progress in developing prosthetic hands in cases of transcarpal amputations, less attention is given to partial-hand and finger prosthetics. Many tasks can still be performed in the absence of one or more fingers but this often comes with limitations and reduced efficiency. Restoring lost fingers through prosthetics could enhance someone's dexterity and precision. Thus, advancing not only prosthetic hands but also prosthetic fingers would enable a greater number of amputees to fully reintegrate into society.

Individuals with partial hand loss can be fitted with either passive silicone cosmeses, body-powered devices, or externally powered devices \cite{imbinto2016treatment}. Each type of prosthesis involves trade-offs between appearance, functionality, and comfort. However, externally powered prostheses have the greatest potential in restoring dexterity and achieving fine motor control, allowing users to perform precise and complex tasks that would be impossible with other prosthetics. 

Nevertheless, few externally powered prostheses are available for the partial hand or fingers. 
Ryu et al. \cite{ryu2020development} proposed a partial hand prosthetic with four pre-programmed grasping patterns that must be manually selected via a switch, and require a preserved thumb and metacarpal stumps. In each mode, the finger motion is triggered by the movement of the thumb.  Murali et al. \cite{murali2019design} designed a 
finger mechanism, attachable to the residual stump, with a length and weight closely resembling that of a natural one. Although an open-loop velocity control system is implemented, no interface is proposed to generate an input based on the user intent.  
To enhance user control over the prosthesis, Kim et al. presented a pneumatically actuated partial prosthetic finger \cite{kim2023development}. Electromyography (EMG) signals, coming from the digitorum flexoris superficialis, were correlated with the air pressure for the pneumatic actuation, though how this correlation is achieved is quite unclear.
These examples highlight a critical gap in powered prostheses that are intuitively controllable, user-friendly, and specifically tailored for individuals with total finger amputation. Indeed, a residual stump, on which to fit the prosthesis, is always required. 

\begin{figure}[t]
    \centering
    \includegraphics[width=\linewidth]{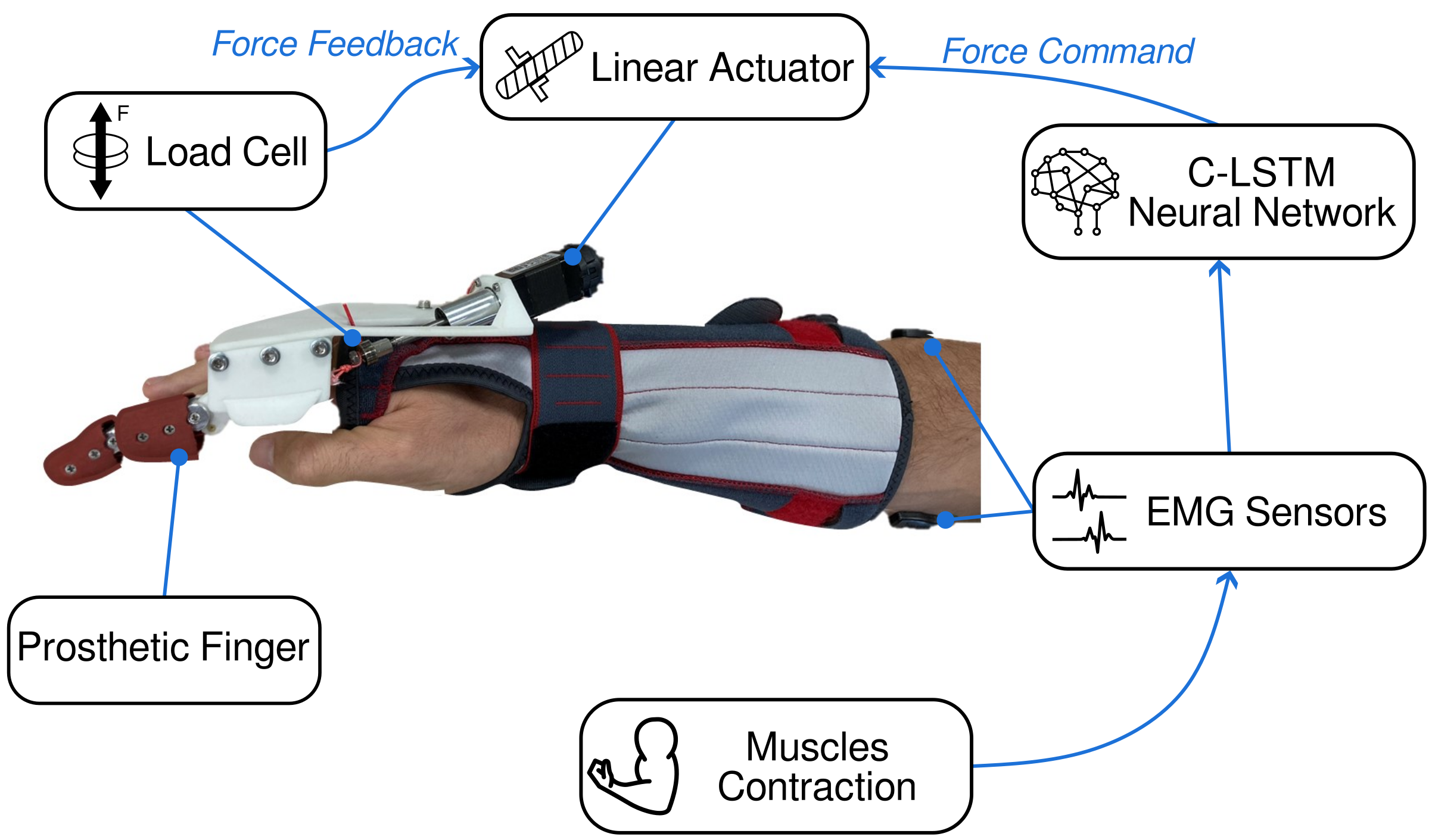}
    \caption{Overview of the main components of the control and actuation scheme of the proposed finger prosthesis}
    \label{fig:intro}
\end{figure}

One possibility to control prostheses intuitively would be to use force control, as muscle forces are the main responsible for joint loading on the physiological level. Yet, the challenge lies in precisely estimating these muscle forces and incorporating them into prosthetic control systems, as direct measurement of muscle forces is invasive, and non-invasive calculations are difficult to perform with high accuracy.
Force-controlled solutions thus rely on electrophysiological signals, like EMG, to dynamically adjust the applied force of the prosthesis. This allows users to handle a wide range of tasks by consciously controlling the forces exerted by the device.
However, the relationship between EMG signals and muscle force is complex and varies significantly across different muscles and individuals. 
For instance, a linear relationship is typically observed in small muscles like the first dorsal interosseous muscle, while larger muscles like the quadriceps femoris exhibit a nonlinear relationship \cite{zhu2017re}.
As a result, computationally expensive methods have been developed to model muscle force using surface EMG inputs. Two primary approaches can be found in literature: neural network(NN)-based models \cite{sitole2023continuous, chen2020cross, zheng2021concurrent, roy2022generic} and decision tree-based models \cite{ameri2014bagged, ye2019optimal}. Both methods offer advantages and are suitable for different estimation tasks, but each has limitations. Crucially, none of these approaches have been extensively tested in controlling actual prosthetic devices.

In this paper, we embark on a first attempt to develop a force-control scheme to intuitively operate a prosthetic finger by leveraging EMG signals (see Fig. \ref{fig:intro}). The device can be controlled by simply contracting the forearm muscles, thus mimicking the natural physiological process. A partial finger prosthetic developed by Airworks was augmented to make it usable by people missing an entire finger. 
We first conducted an experiment with ten unimpaired subjects to identify the most suitable force estimation method for our purpose. The selected model was then used to estimate the fingertip forces online and adjust the tension of the finger artificial tendon, allowing users to actively modulate the grip strength of the device. 
Four subjects wearing the prosthetic finger tested the performance of the developed control architecture, proving the usability of our force estimation model in online applications.


The rest of the paper is organized as follows. Section \ref{sec:models} details the models used to estimate fingertip forces from EMG signals. Section \ref{sec:sysdesc} describes the hardware and low-level control mechanisms. The experimental setup and protocol used to validate the system are outlined in Section \ref{sec:experiments}, and results are presented and discussed in Section \ref{sec:results}.

\section{FINGERTIP FORCE ESTIMATION}
\label{sec:models}
To model the relationship between the fingertip force and EMG signals, three types of regression algorithms were considered: linear regression, ensemble of regression trees, and NN. Scikit-learn \cite{scikit-learn} was used for implementing the first two, while TensorFlow \cite{tensorflow2015-whitepaper} was used for the NN. Of note, we focused on subject-specific models, to address variability caused by individual differences in muscle coordination, tissue filtering, or muscle shape variation during contraction. 


Firstly, we explored the possibility of a linear relationship by computing a linear regression analysis. Then, we implemented two types of ensembled regression trees, Random Forest and Gradient Boosting, using a bagging regressor algorithm. Bagging, also known as Bootstrap Aggregating, improves model robustness and accuracy by training multiple regression trees on different subsets of the training data, created through random sampling with replacement. The final prediction is made by averaging the outputs of all individual trees. This reduces variance and mitigates the risk of overfitting, thus enhancing the model overall performance.
Finally,  we trained a Convolutional Long Short-Term Memory (C-LSTM) NN, with one convolution layer and two LSTM layers. The convolution layer, with 64 filters and a kernel size of 3, was used to extract high-dimensional features from the input data. The two LSTM layers, with hidden sizes of 50 and 30 respectively, were used to capture temporal dependencies, reflecting the relationship between current and previous muscle contraction forces. This architecture leverages the strengths of CNNs and LSTMs to enhance the model ability to interpret complex patterns in the data. The training process was limited to a single epoch with approximately 200 training steps to avoid overfitting.

\section{PROSTHETIC SYSTEM}
\label{sec:sysdesc}
\begin{figure*}[!t]
    \vspace{0.2cm}
	\includegraphics[width=\textwidth]{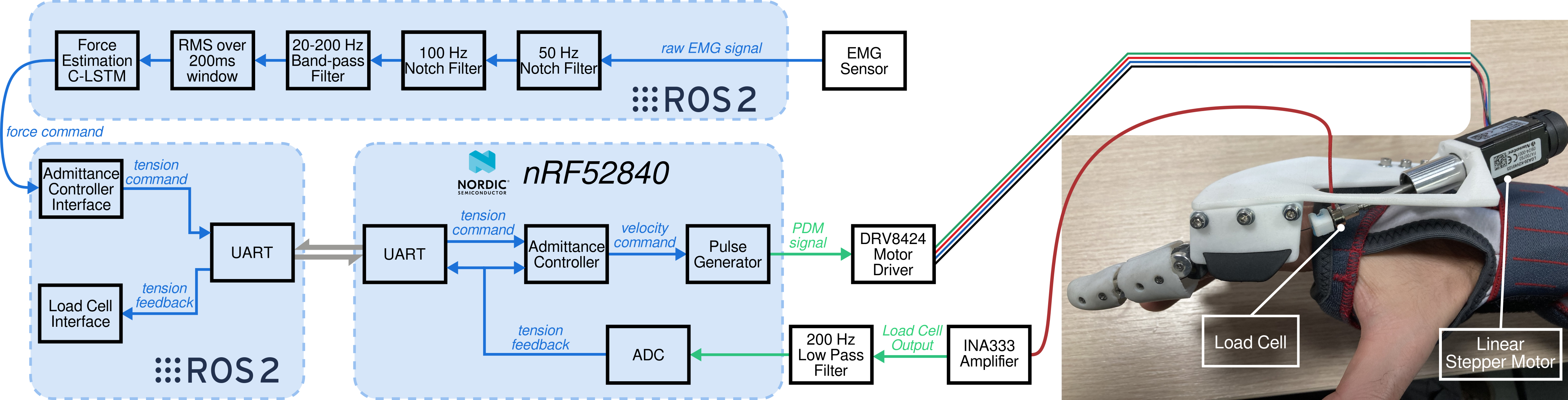}
	\caption{Global schematics of the signal processing and control of the proposed prostetic finger}
	\label{fig:flowchart}
\end{figure*}

 The electronics and control schemes of the proposed prototype are detailed in Fig. \ref{fig:flowchart} and explained in this section.

\subsection{Hardware}
The prosthesis used in this work was originally a body-powered partial finger prosthetic developed by AirWorks, aimed at people missing the distal and intermediate phalanges on one or several fingers, but retaining the ability to move their proximal phalanx. It is a 2-joints underactuated finger, with a single tendon controlling the bending of the finger. Two springs (one for each joint) bring the finger back into its open position when the tendon is not actively pulled. For this study, the device was augmented with a linear actuator to make it usable by people missing an entire finger.

We built a wearable prototype around a wrist brace (Manutonic from Tenortho, ref. TO2217), using the prosthesis as a supernumerary finger placed near the index (see picture in Fig. \ref{fig:flowchart}). Six M2 threaded holes were drilled into the metal plate within the brace, to attach a 3D-printed part holding the prosthesis and the actuator (LGA201S06-B-TBDA-019 from Nanotec). Tension feedback for the closed-loop control is acquired from a load cell (LCM100-FSH04399 from Futek).
The electronics are split into three boards. The main control board is the nRF52840 development kit from Nordic Semiconductors. Two other custom ones are used, respectively for interfacing the load cell and for driving the actuator. The load cell interface board is based on the INA333 instrumentation amplifier from Texas Instruments, and also features a 200 Hz analog low-pass filter. The amplifier gain was set to 212, adjusting the signal to the Analog-to-Digital Converter (ADC) input range. The actuator driver board is based on the DRV8424 stepper motor driver from Texas Instruments.
The prototype weights merely 270g, supported over half of the forearm, and therefore wearing it doesn't cause any discomfort or fatigue.

\subsection{Embedded Software}
The software of the nRF52840 microcontroller relies on the Zephyr embedded operating system as well as some lower-level peripheral control libraries provided by Nordic Semiconductors. It can be divided into four blocks:
\begin{itemize}
    \item The UART block, receiving tension commands from the host computer
    \item The ADC block, reading the load cell output voltage
    \item The main control loop
    \item The pulse generator, generating a Pulse Density Modulated (PDM) signal to control the actuator velocity
\end{itemize}

The system can be used either in open-loop velocity control mode or in closed-loop force control mode thanks to an admittance controller. The latter generates velocity commands from force commands by simulating the motion of a virtual mass-damper system subjected to the desired force. The corresponding equation is:
\begin{equation}
\label{eq:admCtrl}
    m \cdot \dot{v} + d \cdot v = T - T_{cmd},
\end{equation}
where \(m\) is the virtual mass, \(v\) is the velocity, \(d\) is the virtual damping coefficient, \(T\) is the tension of the tendon, measured by the load cell, and \(T_{cmd}\) is the desired tension. Eq. (\ref{eq:admCtrl}) can be rewritten as:
\begin{equation}
    v_{t} = (m \cdot v_{t-1} + \Delta t (T - T_{cmd})) \; / \; (m+d\cdot \Delta t),
\end{equation}
where \(v_{t}\) and \(v_{t-1}\) are the velocities at iterations \(t\) and \(t-1\) respectively, and \(\Delta t\) is the control period. The virtual damping coefficient is set to 1 and the virtual mass to 1 kg. The controller runs at 50 Hz.

\subsection{ROS 2 Interface}
Three ROS nodes are used to control the device. The first one is a low-level UART driver, converting ROS messages into byte arrays and managing the UART communication. The load cell interface only serves logging purposes, and converts the raw output of the microcontroller ADC into force values in Newtons (N). The admittance controller interface purpose is dual, first to convert the fingertip force command obtained from the NN into a tendon tension command, then to make sure the command sent is within the limits of the actuator power. To that end, the command is clamped between 0 and 30 N, and if it increases or decreases by more than 2 N compared to the previous iteration, the command gets smoothed out.

To calculate the tension command from the force command, a two stages model fitting is used. First, the fingertip force, the tension, and the actuator position (related to the bending angle of the finger) are measured. Then, a polynomial surface is fitted to the data, thus obtaining a bidimensional model for calculating the tension from the fingertip force and the actuator position. After this, the latter model is used to plot the tension/fingertip force curves for various constant actuator positions. The second-stage unidimensional model is made by averaging those curves while ensuring monotonicity and that a fingertip force of 0 N would yield a null tension.

While the bidimensional model has better accuracy, its advantage over the unidimensional model is not significant. With the finger being underactuated, the correlation between the joints angles and the actuator position is relatively loose. Therefore, the improvements one can obtain from including the actuator position as a model input are limited. Furthermore, ensuring the aforementioned desirable properties is easier to achieve with an unidimensional model, and the computational cost is lower.

\section{EXPERIMENTS}
\label{sec:experiments}
The proposed system was evaluated with two experiments. First, the force estimation model was validated offline on ten subjects. Next, four participants wearing the prosthesis tested the performance of the online EMG-based control developed.

\subsection{Signal Processing}
\label{sec:sp}
The EMG signals were processed with SciPy \cite{2020SciPy-NMeth}. The data from each EMG sensor was filtered through a 4th-order Butterworth filter with a 20-200Hz passband, and a series of notch filters with rejection frequencies at multiples of 50Hz were applied to remove power line interference \cite{chen2020cross}. Then, the Root of the Mean of Squares (RMS) of the filtered data was computed over a 500ms window. Finally, the RMS values and the F/T sensor signal were normalized with respect to their maximum values.

\subsection{Experimental setup and protocol}
\begin{figure}[t]
\vspace{0.2cm}
  \centering
  \subfloat[Flexor]{\includegraphics[width=0.45\textwidth]{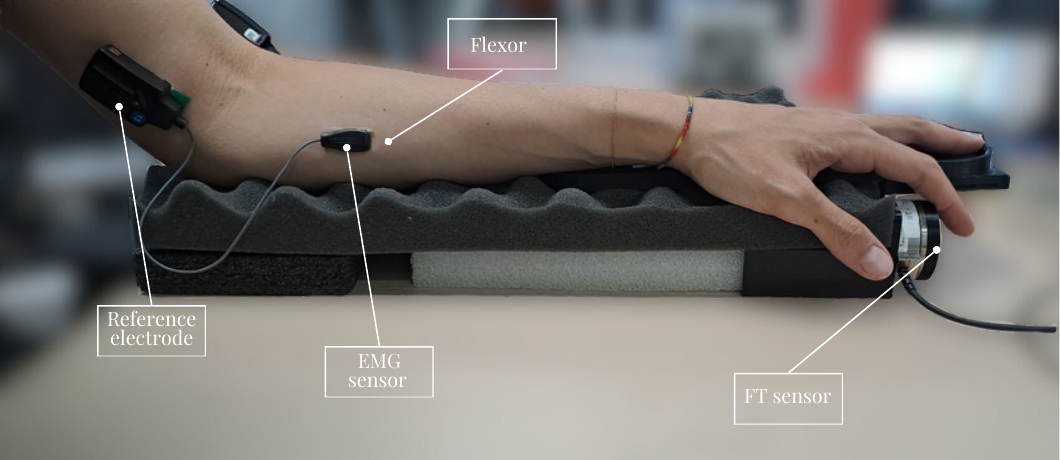}} \\
  \subfloat[Extensor]{\includegraphics[width=0.45\textwidth]{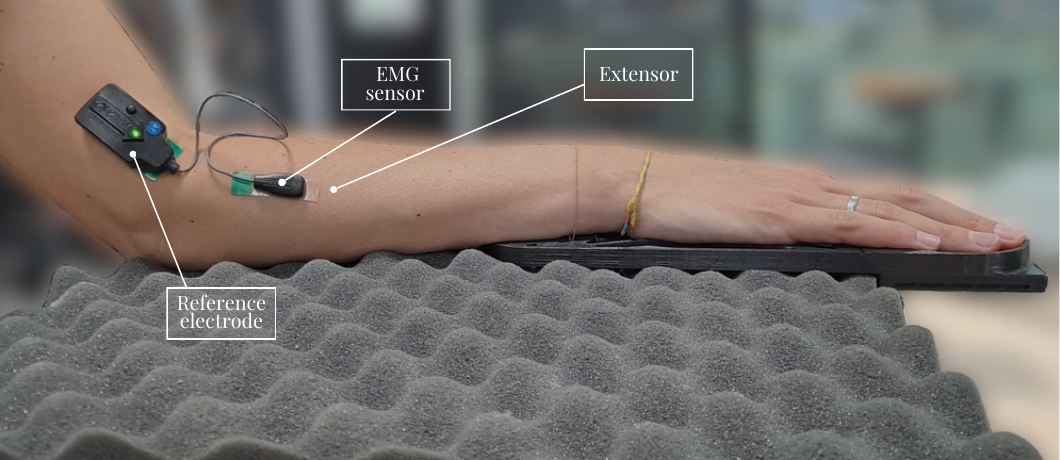}}
  \caption{Setup and sensors employed in the experiments}
  \label{fig:setup}
\end{figure}

Fig.~\ref{fig:setup} depicts the setup and sensors employed in the experiments. We used a Force/Torque (F/T) sensor (Mini45 from ATI) to measure the fingertip force, and a Trigno$^\text{\textregistered}$ Research+ System from Delsys (Delsys Incorporated, Natick, MA, USA) to collect the EMG signals.
Two Trigno mini EMG sensors were placed on the contralateral flexor digitorum superficialis and on the extensor digitorum superficialis, respectively, as it is known that finger flexion involves co-contraction of these forearm muscles. The reference electrodes were placed at the elbow level. Before placing the electrodes, the subjects’ forearms were shaved, and the skin surface was cleaned with alcohol pads.

\subsubsection{Force Estimation Experiment}
\label{sec:data_collection}
Ten unimpaired volunteers (five female, five male), participated in the experiment. Subjects were asked to sit comfortably on a chair and to place their left forearm on a soft support fixed to the table, with the palm parallel to the table surface and the fingers extended.
This positioning of the forearm was selected to minimize interference from other muscles in the forearm. Subjects were then asked to press with their index finger on the F/T sensor fixed to the table.
In the first stage of the experiment, the Maximum Voluntary Contraction (MVC) of the index finger was measured over 30 seconds during maximum isometric flexion, and the maximum RMS value from the filtered EMG signals as well as the maximum force exerted on the F/T sensor were recorded. In the second stage, subjects were asked to follow a pseudo-randomly generated force pattern, varying between 25\%, 50\%, and 60\% of the previously recorded maximum force, for a duration of 250 seconds. Visual feedback showing the force pattern and the current force was provided. Each subject performed the experiment twice, with 5 minutes breaks between trials. One trial was used for training the force estimation models, and one for evaluation.

\subsubsection{Prosthesis Control Experiment}
The setup depicted in Fig.\ref{fig:setup} was used, but with the prosthetic finger pressing on the F/T sensor instead of the user's own finger. Four subjects out of ten (two female, two male), for whom the wearable prototype fit well, took part in this second experiment. They were asked to wear the prosthesis and follow a force pattern as in the previous experiment. However, some modifications were made to ease the control of the prosthesis. First, there was a perceptible delay between the muscle contraction and the rise of the generated force command, mainly due to the 500ms sliding window used to compute the RMS. Therefore, we reduced this window to 200ms, making the delay imperceptible. Then, the output of the force estimation model was scaled to [0-8] N instead of being scaled up to each individual maximum force; the 60\% of the maximum force used as reference to generate the force pattern then amounted to 4.8N, close to the limit of what the actuator can deliver. Lastly, reaching null force requires complete relaxation from the subjects, which proved to be difficult to achieve for most people. Therefore, the force commands below 1N were brought down to 0N.

It is noteworthy that while the force estimation models were trained on data from the left arm, this experiment was performed with EMG data from the right arm. Since force cannot be measured from the absent limb in amputees, the forces were acquired from the contralateral limb, simulating the absence of the right index finger. This strategy is supported by the observation that a significant correlation exists between the right and left upper limb forces during mirrored contractions \cite{oda1997motor}. Thus, obtaining a model from an intact limb does not introduce significant variation in the results.

\subsection{Experimental Analysis}

\begin{figure}[t]
\vspace{0.2cm}
  \centering
  \subfloat{\includegraphics[width=\linewidth]{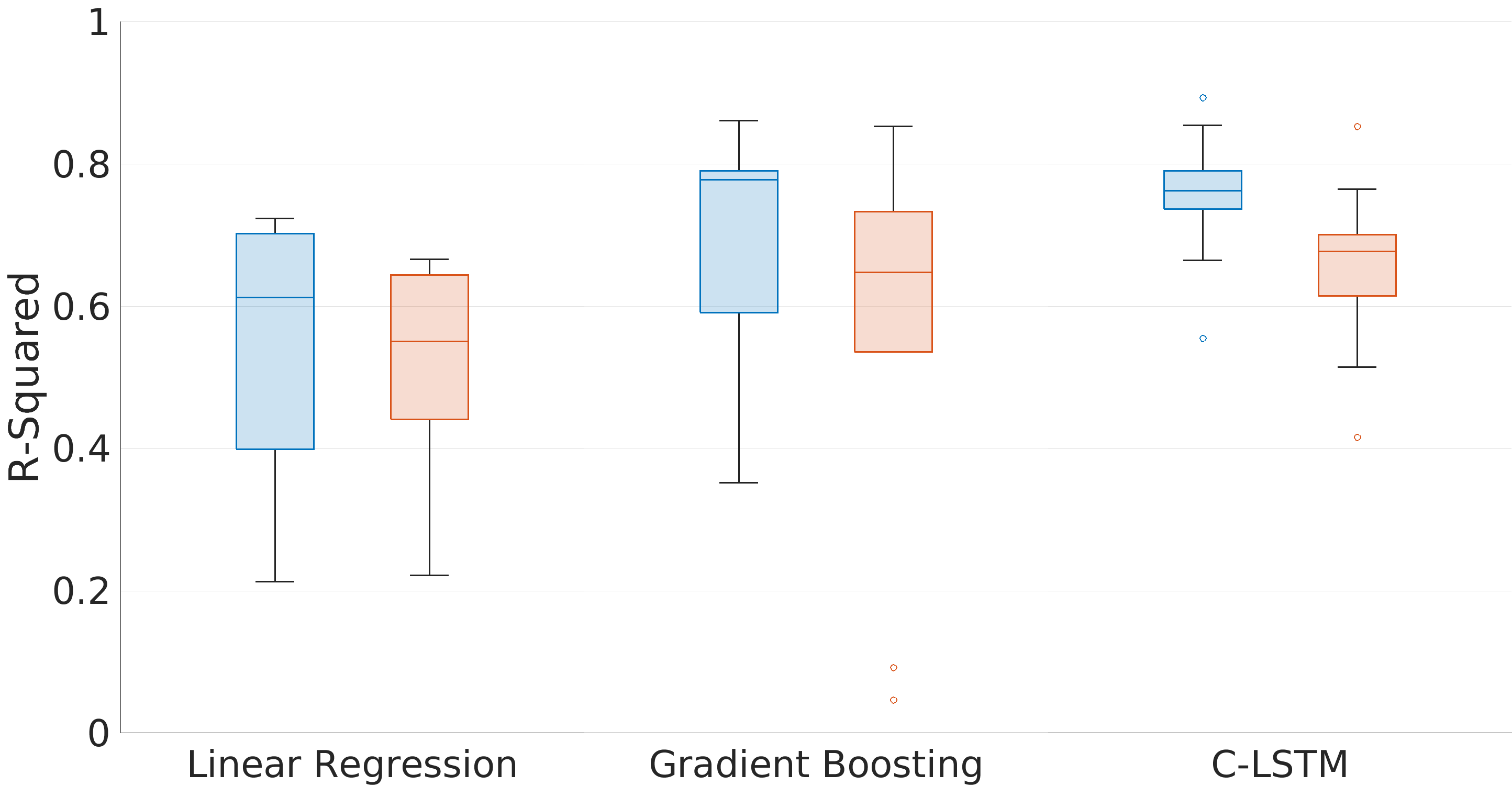}} \\
  \subfloat{\includegraphics[width=\linewidth]{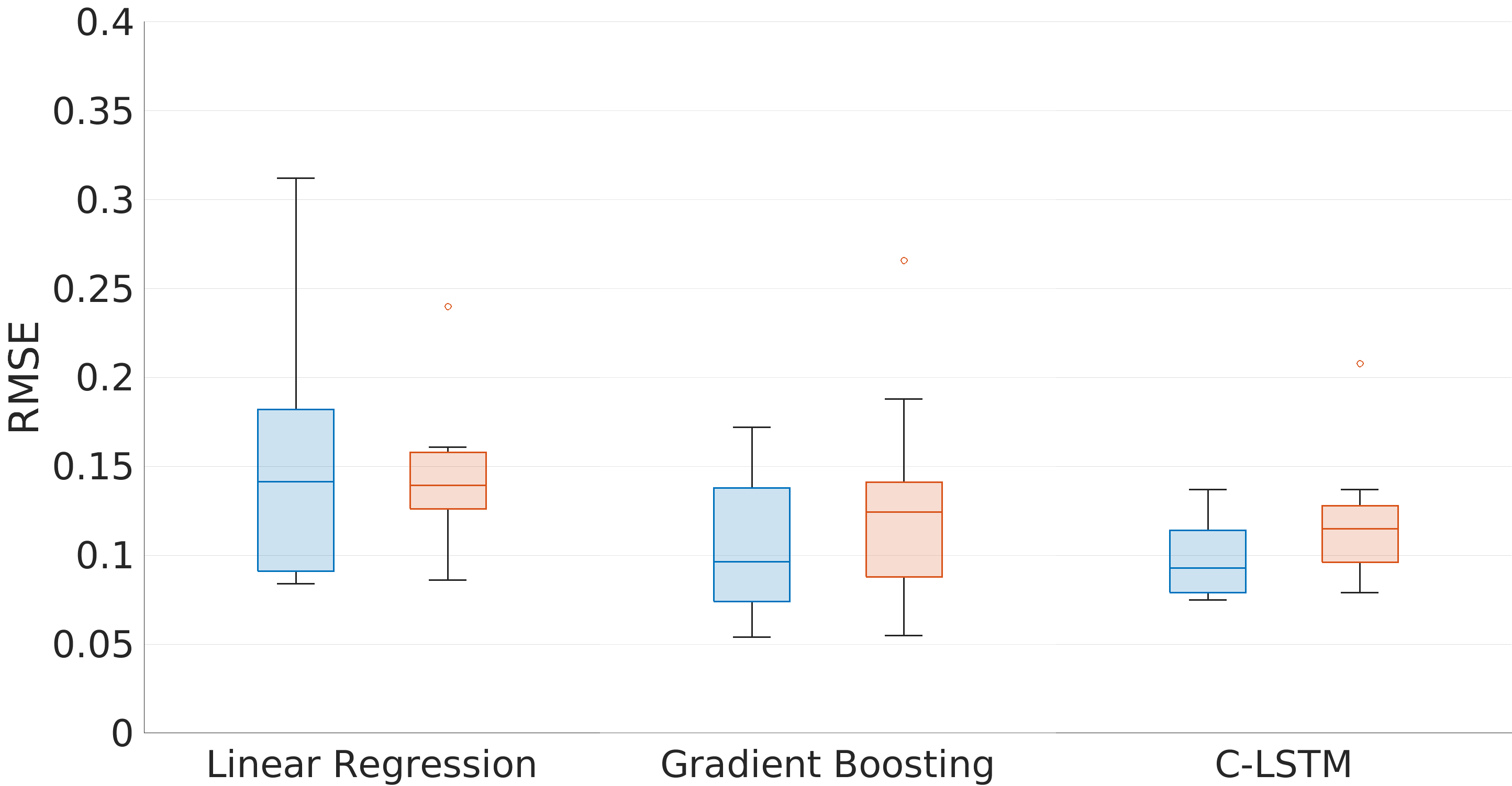}} \\
  \subfloat{\includegraphics[width=0.3\linewidth]{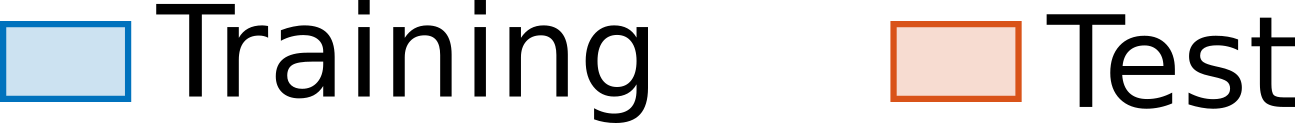}}
  \caption{Statistical performance of each force estimation method, on training and test datasets, across all ten subjects}
  \label{fig:results_NN}
\end{figure}

\subsubsection{Force Estimation Experiment}
The collected data were used to test the force estimation approaches found in the literature and mentioned in Section \ref{sec:models}: linear regression, Random Forest, Gradient Boosting, and C-LSTM. The outputs of these methods were compared to determine the most effective force estimation model for our application. 

For this analysis, we used two different performance metrics, both on training and test data: the coefficient of determination (R-squared) and the Root Mean Squares Error (RMSE) \cite{tyagi2022regression, zheng2021concurrent}. The R-squared value indicates how well the model explains the variance in the data, with higher values indicating a better fit. The RMSE measures the average magnitude of the prediction errors, providing insight into the model accuracy (i.e., it tells how close the predicted values are to the actual values).

\subsubsection{Prosthesis Control Experiment}
The following characteristics of the device were tested during the experiment: the controllability of the device, the admittance controller performance, and the accuracy of the model to obtain the tension of the tendon from the desired fingertip force.

Three aspects of the controllability of the device were evaluated. First, the system capability to track a given force command, which we called force tracking. Second, the ability of the user to generate a force command following the given force pattern by contracting or relaxing its forearm muscles, which we called force targeting. Lastly, combining the latter two, the ability of the user to apply a given force level on the F/T sensor with the prosthesis, which we called force reaching. As performance metrics to measure these three aspects, we calculated the RMSE between the force command and the force applied onto the F/T sensor, between the target force pattern and the generated force command, and between the target force pattern and the force applied onto the F/T sensor, respectively.

The admittance controller performance can be verified by looking at the graph showing the tension command and the tension feedback provided by the load cell in Fig. \ref{fig:tension_tracking}. 

The accuracy of the model converting the fingertip force into tendon tension was measured by the RMSE and R-squared values between the fingertip force measured by the F/T sensor and the fingertip force estimated from the model based on the corresponding tension value.

\section{RESULTS AND DISCUSSION}
\label{sec:results}

\begin{figure}[t]
\vspace{0.2cm}
    \centering
    \includegraphics[width=\linewidth]{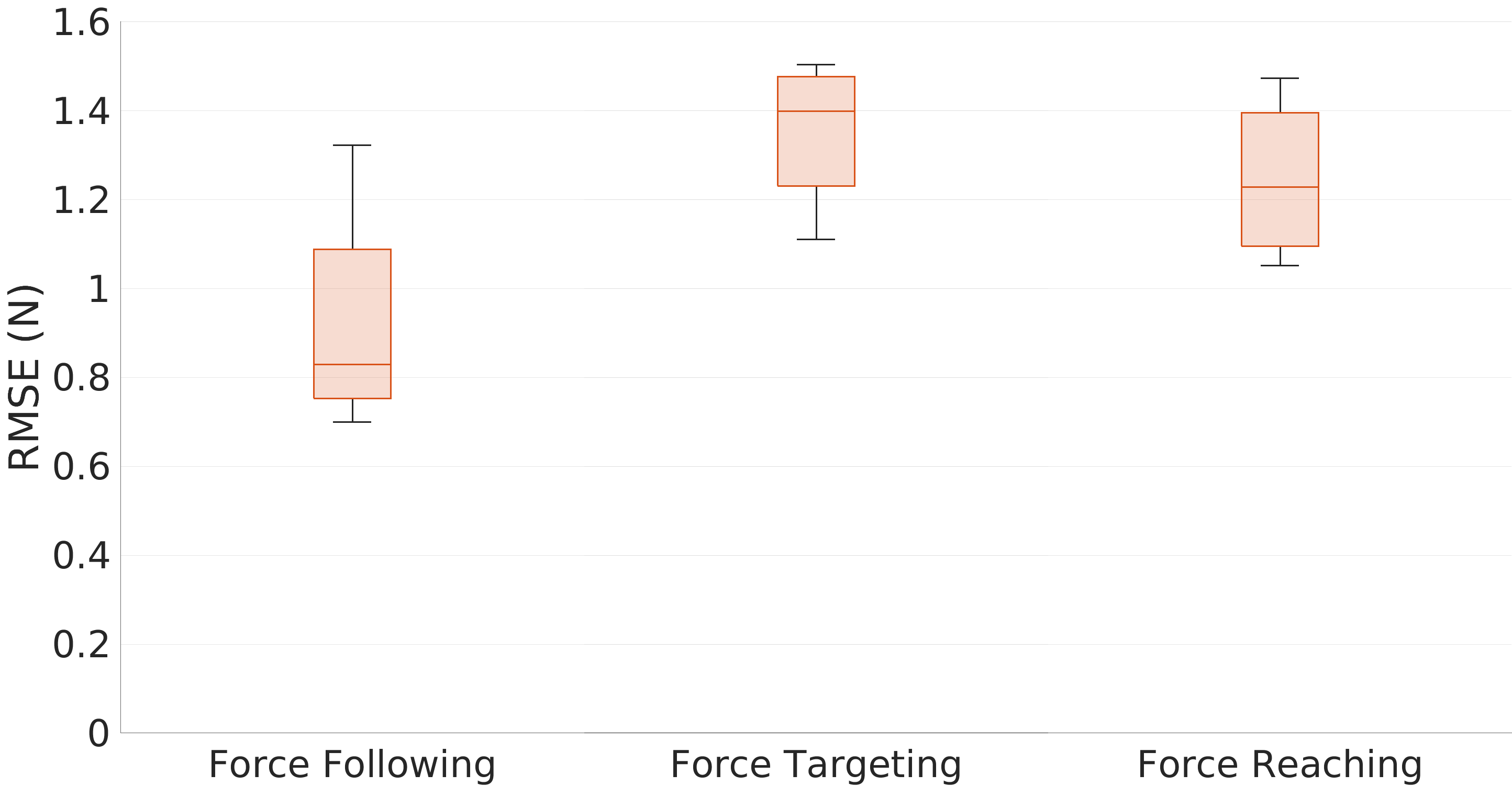}
    \caption{RMS error statistics across all four participants during the prosthesis control experiments, between the command and the output (force following), between the pattern and the command (force targeting), and between the pattern and the output (force reaching)}
    \label{fig:stats}
\end{figure}

\subsection{Validation of the Force Estimation Model}
The performance results of the tested methods are reported in Fig.~\ref{fig:results_NN}.
The linear regression analysis yielded poor results, even on the training set, demonstrating that a linear relationship is not sufficient to correlate the EMG signals of the flexor and extensor digitorum superficialis muscles and fingertip forces. The Random Forest approach delivered the lowest results, producing negative R-squared values in some subjects on the test set. For this reason, the method was discarded and its results are not shown in Fig.~\ref{fig:results_NN}. Both Gradient Boosting and C-LSTM produced stronger training-set R-squared and RMSE values compared to Linear Regression. However, the Gradient Boosting model tended to show a more significative drop in performance on the test set, and in some cases, its test R-squared values were substantially lower than those of the C-LSTM, as depicted by the two outliers in Fig.~\ref{fig:results_NN}.
In contrast, the C-LSTM performance on the test set was occasionally even better with respect to the training set, with the highest average correlation coefficient and the lowest average RMSE. Moreover, the C-LSTM exhibited the least performance degradation between the training and test phases, for both R-squared and RMSE metrics, showing that the model has good generalization performance.

\begin{figure}[htbp]
    \vspace{0.2cm}
    \centering
    \subfloat[\label{fig:a}]{\includegraphics[width=\linewidth]{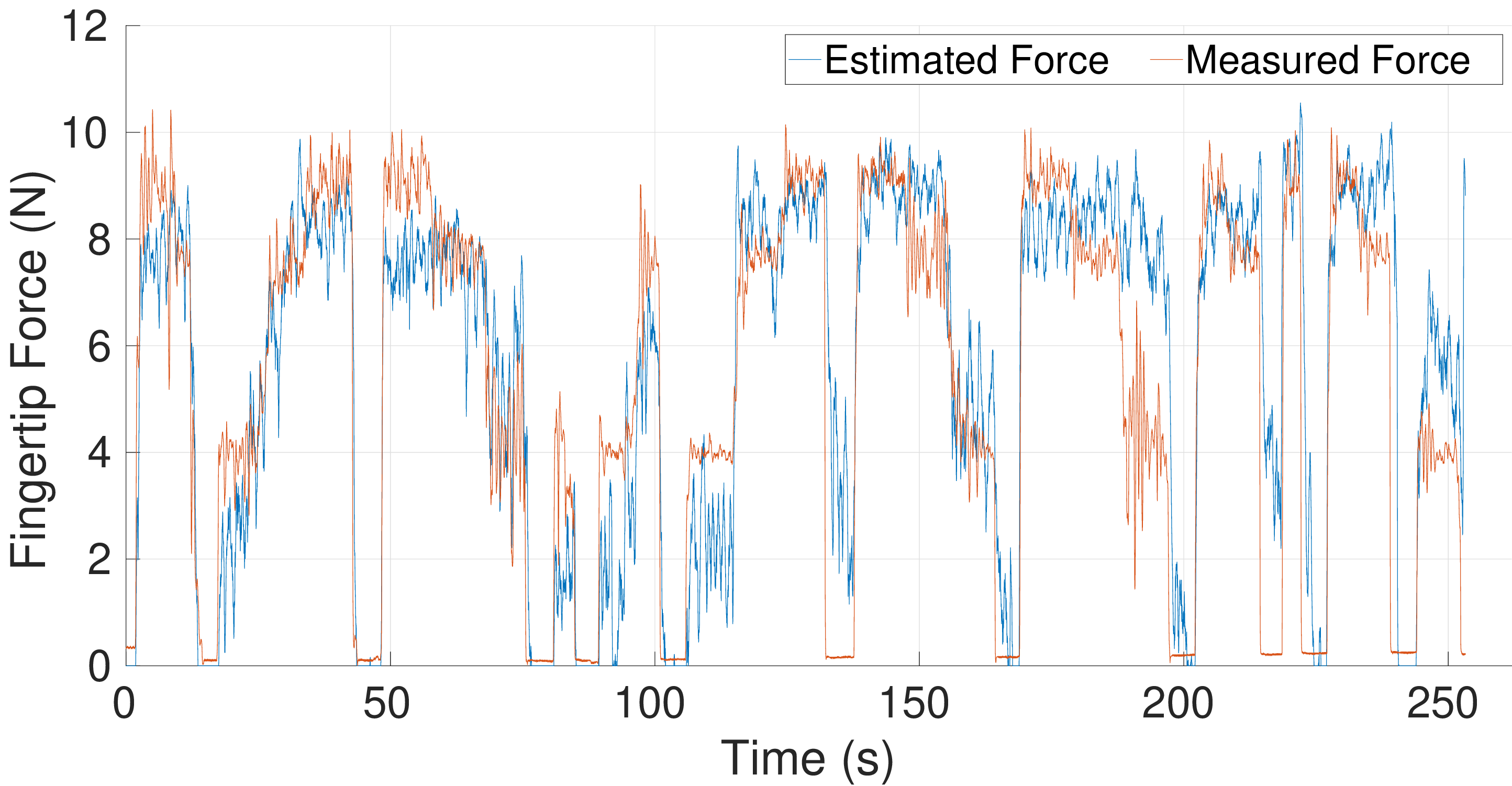}}
    
    \subfloat[\label{fig:force_tracking}]{\includegraphics[width=\linewidth]{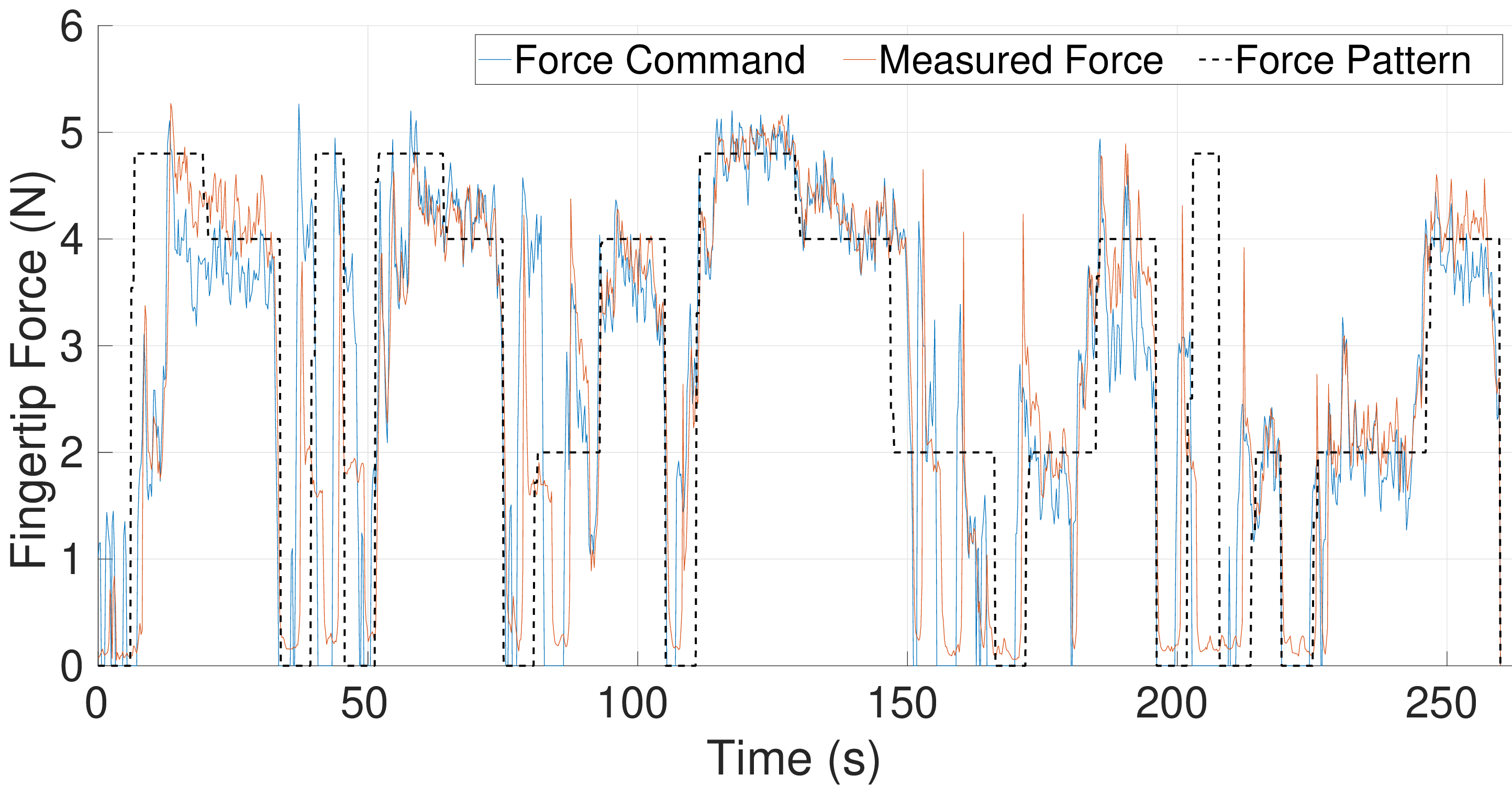}}
    
    \subfloat[\label{fig:tension_tracking}]{\includegraphics[width=\linewidth]{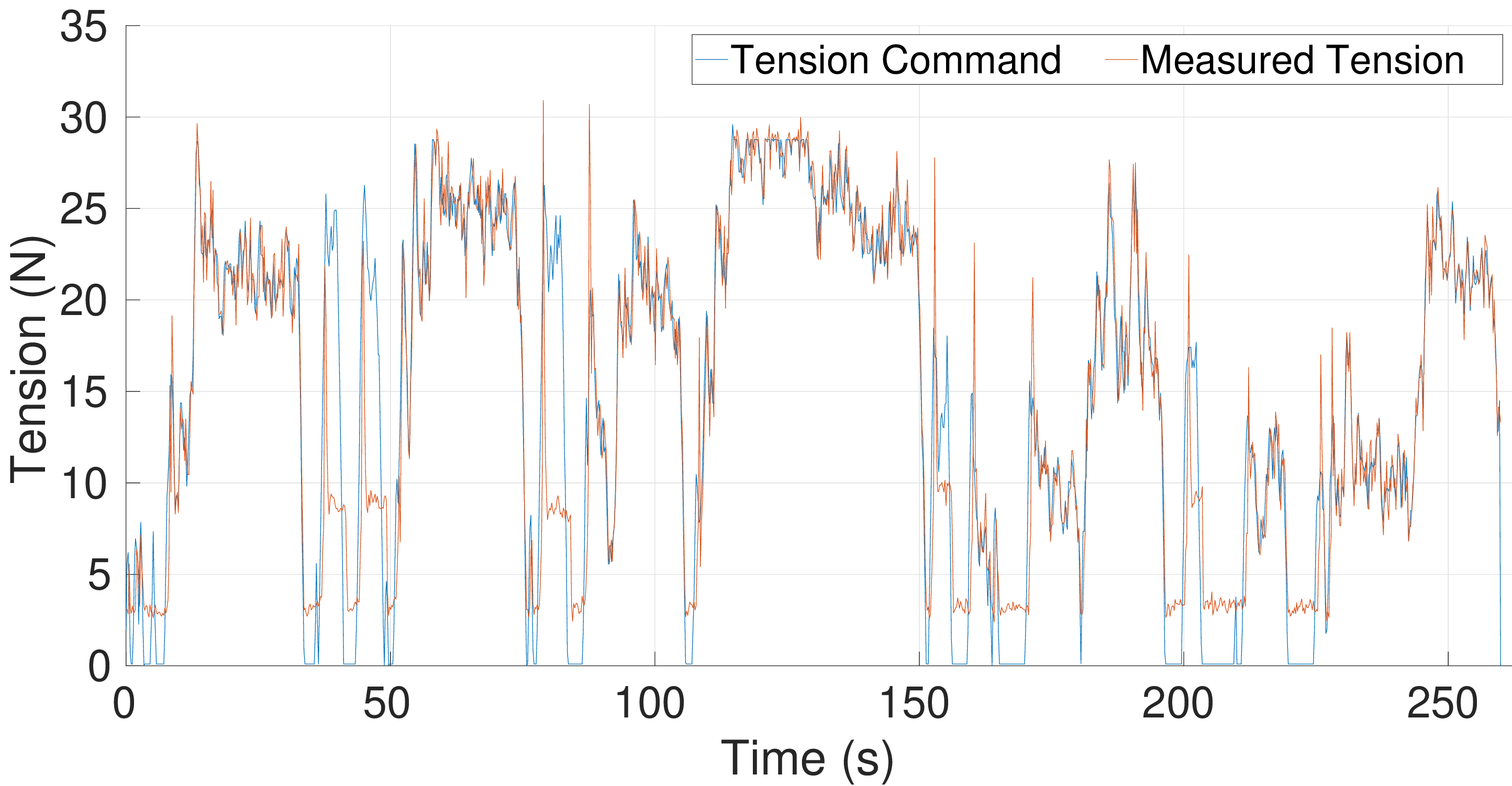}}

    \subfloat[\label{fig:force_tension_model}]{\includegraphics[width=\linewidth]{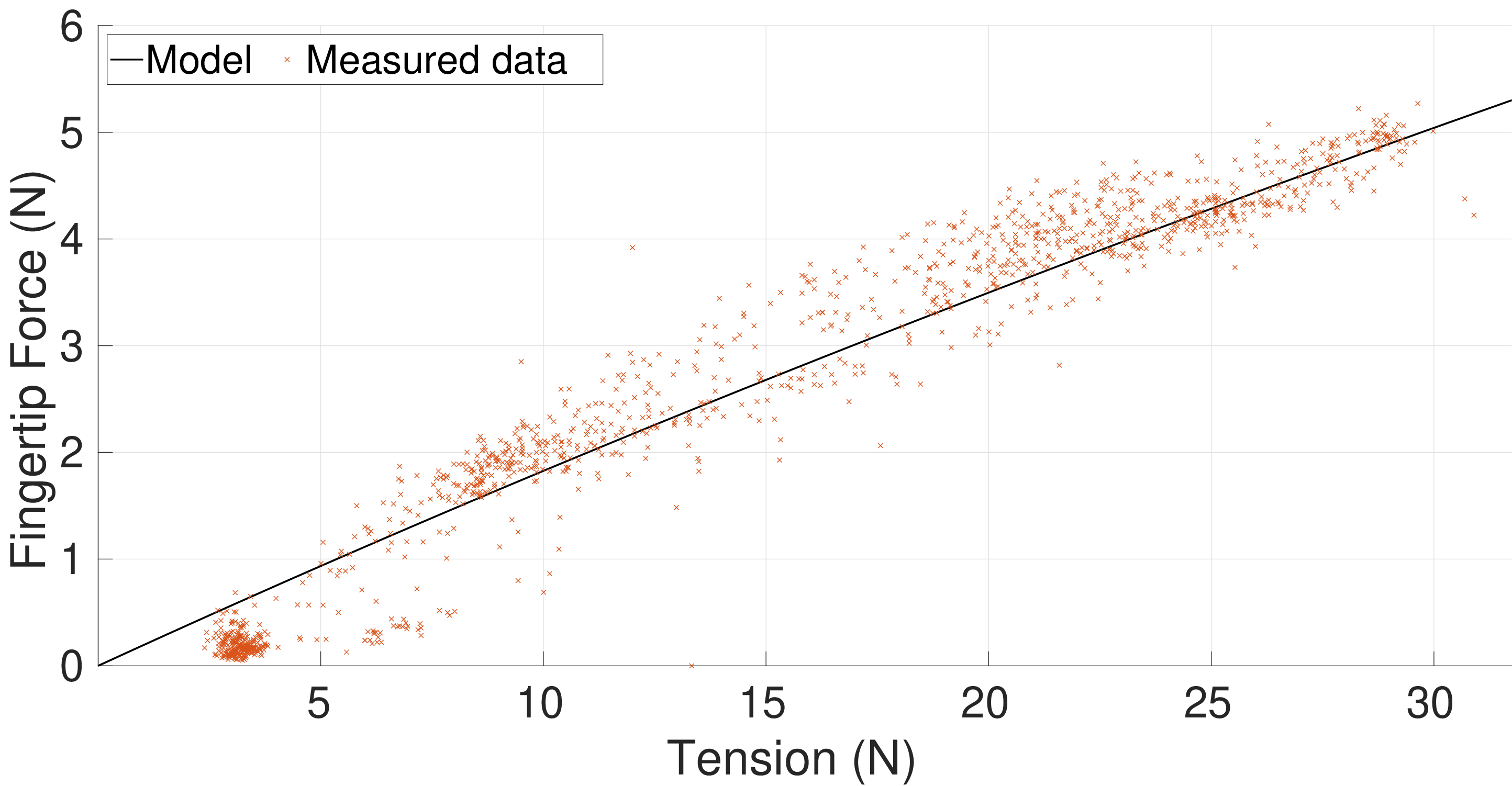}}
    
    \caption{Experimental results for subject 2 \textbf{(a)} Comparison of the force estimated by the C-LSTM and the actual force applied by the user \textbf{(b)} Comparison between the force command generated by muscle contraction, the force applied by the prosthetic system, and the target force \textbf{(c)} Tension tracking \textbf{(d)} Model used for calculating the tension command depending on the force command}
    \label{fig:plots}
\end{figure}

This led us to select the C-LSTM model as the best method to establish the relationship between EMG signals and fingertip forces.
Fig.~\ref{fig:a} shows the plot of the measured force recorded by the F/T sensor versus the force estimated by the C-LSTM for subject 2. It is noticeable how the two forces have a similar trend throughout the experiment, indicating that the estimated force closely tracks the measured force, with minor deviations.

\subsection{Evaluation of Prosthesis Controllability}
Statistical results of the prosthesis control experiment are presented in Fig. \ref{fig:stats}. The force tracking results show that the hardware offers satisfying performance, with RMSE lower than 1 N on average. Fig. \ref{fig:force_tracking} also shows that the force applied to the F/T sensor consistently follows the command. The force targeting and force reaching metrics have a slightly higher RMSE but lower than 1.5 N, probably due to the influence of human factors.
The slightly lower value in force reaching compared to the force targeting shows that users adapted to the system by adjusting the position of their arm to help them follow the pattern when they had difficulties adjusting their muscle contraction to the right amount. This adaptability of the prosthesis user removes the need for a very high accuracy of the system, since little discrepancies are intuitively compensated.


Fig. \ref{fig:tension_tracking} depicts the tension command and the load cell tension feedback, providing clear evidence of the two signals similarity. 
The fact that the measured tension doesn't go down all the way to zero is due to the springs in the prosthesis joints, and to the user maintaining contact between the prosthetic finger and the F/T sensor (as can be seen in Fig. \ref{fig:force_tracking}, the force applied onto the sensor doesn't drop all the way to zero either). 

The relationship between the tendon tension and the fingertip force shows some significant variability, as can be seen in Fig. \ref{fig:force_tension_model}. Indeed, the fingertip force depends on several factors including the joints angles of the finger and the position of the contact point, neither of those being observable with our system.
Per-subject results are shown in Table \ref{tab:force_tension_model}. The variations observed between the subjects can be related to different configurations of the arm. 

\begin{table}[t]
\vspace{0.2cm}
\centering
$\begin{array}{@{}cc cc cc@{}}
\toprule
\text{subject n°} & \text{RMSE} & \text{R-squared} \\
\cmidrule[0.4pt](r{0.125em}){1-1} \cmidrule[0.4pt](lr{0.125em}){2-2} \cmidrule[0.4pt](lr{0.125em}){3-3}
1 & 1.17 & 0.928  \\ 
2 & 0.387 & 0.976 \\ 
5 & 0.508 & 0.939 \\ 
8 & 0.772 & 0.942 \\ 
\midrule
\textbf{Average} & \textbf{0.710} & \textbf{0.946} \\
\bottomrule
\end{array}$
\caption{Fingertip force to tension model accuracy}
\label{tab:force_tension_model}
\end{table}

\section{CONCLUSION}
In this work, we presented a force-controlled prosthetic finger using EMG signals and a C-LSTM NN, enabling users to intuitively control the device. 
Our model demonstrated accurate force estimation, allowing users to adjust grip strength dynamically, with performances consistent across different subjects and good generalization to unseen data.


The EMG-based force control of the prosthetic proved to be intuitive, and subjects were able to use the system without extensive training. However, they found it difficult to completely relax their forearms, which is the only way the grip force of the finger can be brought to zero. A wider range of positions and movements during the data acquisition for training the model might allow to release the finger grip by contracting the extensor muscles in the future, which would boost the intuitiveness of our control scheme further. 
The admittance controller showed very satisfying performance, demonstrating that using actuator without direct force control capability is not an obstacle to build force-controlled prosthetics. Finally, it was shown that the EMG-based force control can tolerate little inaccuracies for the application to finger prostheses, as the users tend to intuitively compensate for these. All these points make the EMG-based force control a valid proof of concept for upper limb prosthetics.

\addtolength{\textheight}{-10cm}   




\section*{ACKNOWLEDGMENTS}
The authors would like to thank Alberto Parmiggiani for his advice on the mechanical design of the wearable prototype; Davide Dellepiane for manufacturing the motor driver PCB; and Phil Edward Hudson and Marco Migliorini for providing the load cell interface PCB.

\bibliographystyle{IEEEtran}
\bibliography{bibliography.bib}
\end{document}